\documentclass[conference]{IEEEtran}
\IEEEoverridecommandlockouts
\usepackage{cite}
\usepackage{amsmath,amssymb,amsfonts}
\usepackage{algorithmic}
\usepackage{graphicx}
\usepackage{textcomp}
\usepackage{xcolor}
\usepackage{url}
\usepackage{hyperref}
\hypersetup{
    colorlinks=true,
    linkcolor=blue,
    filecolor=magenta,      
    urlcolor=cyan,
}

\def\BibTeX{{\rm B\kern-.05em{\sc i\kern-.025em b}\kern-.08em
    T\kern-.1667em\lower.7ex\hbox{E}\kern-.125emX}}
\begin{document}

\title{Bringing A Robot Simulator to the \\ SCAMP Vision System
\thanks{This work is accepted by ICRA2021 workshop On and Near-sensor Vision Processing, from Photons to Applications (ONSVP). This work was supported by UK EPSRC EP/M019454/1, EP/M019284/1, EPSRC Centre for Doctoral Training in Future Autonomous and Robotic Systems: FARSCOPE and China
Scholarship Council (No. 201700260083).}
}

\makeatletter
\newcommand{\linebreakand}{%
  \end{@IEEEauthorhalign}
  \hfill\mbox{}\par
  \mbox{}\hfill\begin{@IEEEauthorhalign}
}
\makeatother

\author{
\IEEEauthorblockN{Yanan Liu}
\IEEEauthorblockA{\textit{Bristol Robotics Laboratory} \\
\textit{University of Bristol}\\
Bristol, UK \\
yanan.liu@bristol.ac.uk}
\and
\IEEEauthorblockN{Jianing Chen}
\IEEEauthorblockA{\textit{School of Electrical \& Electronic Engineering} \\
\textit{University of Manchester}\\
Manchester, UK \\
jianing.chen@manchester.ac.uk}
\and
\IEEEauthorblockN{Laurie Bose}
\IEEEauthorblockA{\textit{Visual Information Labratory} \\
\textit{University of Bristol}\\
Bristol, UK \\
lauriebose@gmail.com}
\linebreakand
\IEEEauthorblockN{Piotr Dudek}
\IEEEauthorblockA{\textit{School of Electrical \& Electronic Engineering} \\
\textit{University of Manchester}\\
Manchester, UK \\
p.dudek@manchester.ac.uk}
\and
\IEEEauthorblockN{Walterio Mayol-Cuevas}
\IEEEauthorblockA{\textit{Visual Information Laboratory} \\
\textit{University of Bristol}\\
Bristol, UK \\
walterio.mayol-cuevas@bristol.ac.uk}
}

\maketitle

\begin{abstract}
This work develops and demonstrates the integration of the SCAMP-5d vision system into the CoppeliaSim robot simulator, creating a semi-simulated environment. By configuring a camera in the simulator and setting up communication with the SCAMP python host through remote API, sensor images from the simulator can be transferred to the SCAMP vision sensor, where on sensor image processing such as CNN inference can be performed. SCAMP output is then fed back into CoppeliaSim. This proposed platform integration enables rapid prototyping validations of SCAMP algorithms for robotic systems. We demonstrate a car localisation and tracking task using this proposed semi-simulated platform, with a CNN inference on SCAMP to command the motion of a robot. We made this platform available online.

\end{abstract}


\section{Introduction}

The SCAMP visual system is a smart sensor supporting in-sensor processing. The direct analogue electronic current signal process and calculation on the pixel processor array (PPA) enable low-power consumption, parallel, and efficient computing without external hardware. With these features, it is increasingly being integrated with robots for various applications \cite{9197370,https://doi.org/10.1049/ipr2.12158,8206286,mcconville2020visual,weighted2021}. However, it is often time-consuming and difficult to prototype ideas using real robotic platforms, especially during the COVID-19 pandemic period. To improve experimental flexibility, we integrate a comprehensive robot simulator CoppeliaSim\cite{rohmer2013v} and SCAMP python host to test and validate ideas rapidly. CoppeliaSim is a robot environment simulator where each agent can be controlled via remote API \cite{james2019pyrep}. Its simulated sensor readings can be transferred to other independent platforms written in python, C/C++, or Matlab through several communication protocols. Based on the proposed semi-simulation platform, we implemented a convolutional neural network (CNN) \cite{liu2020bmvc,bose2020} on the SCAMP processing the imported camera images for localisation purpose from the robot simulator where the camera is mounted under a drone.

\section{SCAMP Python Host and the Semi-Simulated Platform}

\begin{figure}[t]
\centering
\includegraphics[width=2.5in]{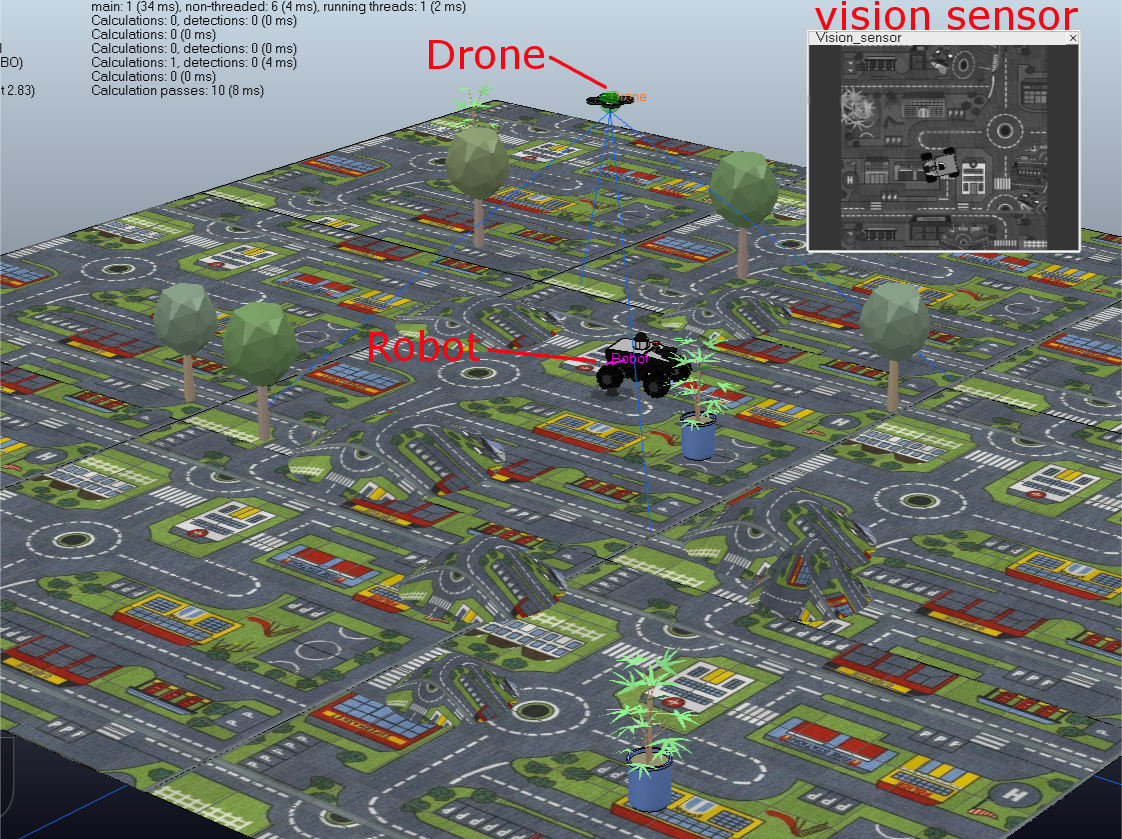}
\caption{Robot simulation environment. The SCAMP is 'mounted' under the drone facing the ground. Real-time image can be seen from the vision sensor with a resolution of 256$\times$256 which is set the same as that of a SCAMP. The floor is designed with disrupting texture for CNN localisation. Note that the current version of SCAMP-5d only supports gray-scale images; hence CNN inference on SCAMP only relies on gray texture from the scene.}
\label{fig:localisation}
\end{figure}

A SCAMP vision system is connected to a computer via USB through scamp5d\_interface library. And SCAMP host \cite{chen2018scamp5d} \footnote{\url{https://scamp.gitlab.io/scamp5d_doc/_p_a_g_e__g_u_i_d_e__g_u_i.html}} is a GUI executed on the computer to interact with the vision system and visualise the data sent back from the device. This work develops a scamp\_python\_module for the python GUI based on previous C/C++ host libraries, to make the SCAMP host easier to connect to third party software. With this method, the host visualisation and remote API can be co-designed on the scamp\_python\_module. The configuration of remote API on the CoppeliaSim can be found from \footnote{\url{https://www.coppeliarobotics.com/helpFiles/en/b0RemoteApiOverview.htm}}.

\begin{figure}[t]
\centering
\includegraphics[width=3.2in]{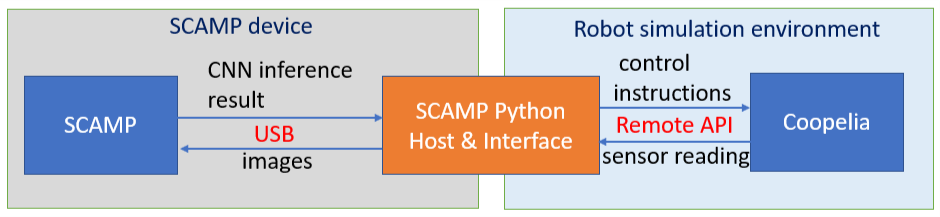}
\caption{Semi-simulated platform by integrating the SCAMP with Coppeliasim Robot Simulator. This platform takes advantage of the SCAMP parallel computation ability and the rich simulation scenes in the simulator, where applications can be exploited and validated virtually.}
\label{fig:platform}
\end{figure}

To evaluate the CNN performance on SCAMP working with a robot-related applications, this paper developed a semi-simulated platform in which a real SCAMP-5d visual system collaborates with Coppeliasim robotics simulation environment through remote API\footnote{\url{https://www.coppeliarobotics.com/helpFiles/en/blueZeroPlugin.htm}}. That means environment setups and sensor image collection is performed in the simulator while the real SCAMP is in charge of CNN inference with sensor images from the simulator and outputting useful information to the simulator (Fig. \ref{fig:platform}).

\section{Experiments on platform}

\begin{figure}[t]
\centering
\includegraphics[width=3.0in]{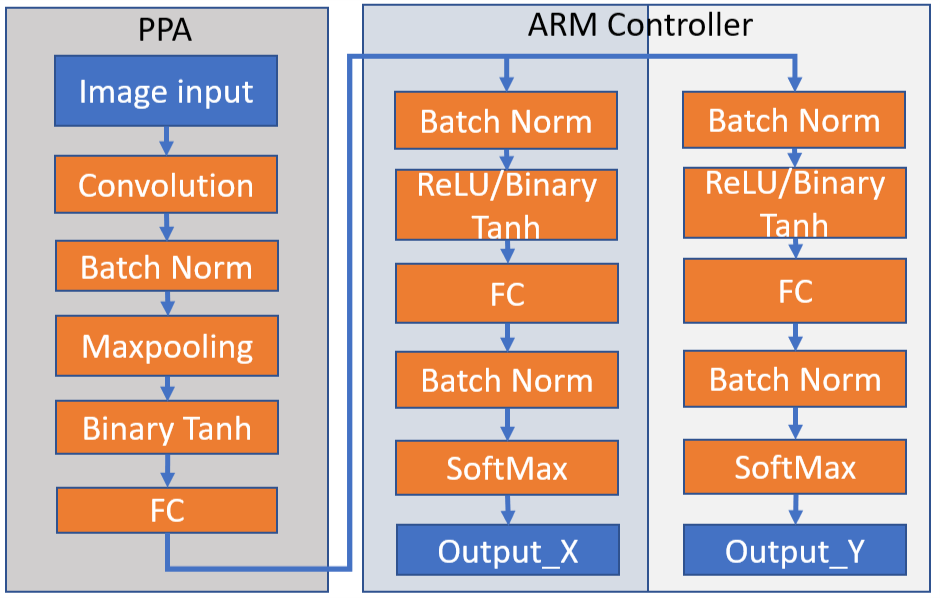}
\caption{The CNN architecture for car localisation. A shared convolutional layer is used for object 2D localisation on the SCAMP, where label \textit{x} and \textit{y} using identical convolutional layer but different fully-connected layers.}
\label{fig:cnn-localisation}
\end{figure}

\subsection{binarized CNN for car localisation from a drone}

This work trained a binarized CNN which introduces batch norm and use both binary weights and activations to reduce the error caused by continuous analogue electronic current computing \cite{zhou2020near}. A car localisation and tracking task are exploited based on the proposed neural network structure. SCAMP visual system is suitable for mobile robot platform due to its low power consumption, lightweight and in-sensor machine vision computing ability without using or communicating with external hardware. This experiment takes advantage of the SCAMP-5d visual system to localise a mobile vehicle moving in the 2D simulated environment, using this location information to guide a drone to track the vehicle.

\subsubsection{Localisation Dataset}

\begin{figure}[t]
\centering
\includegraphics[width=3.0in]{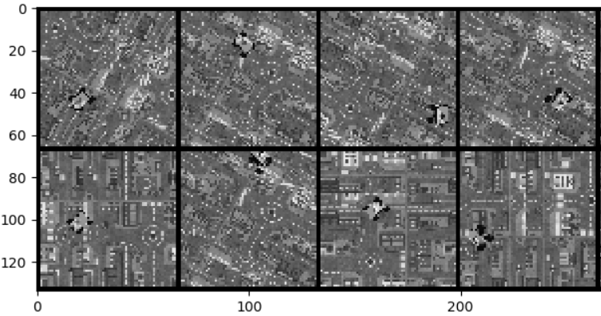}
\caption{Selected images for training. Although from the simulation environment the localisation seems simple, the inputs for SCAMP is the gray-scale images with a low resolution of 64$\times$64, which is a challenging task for SCAMP localisation.}
\label{fig:car_dataset}
\end{figure}

The localisation training dataset is collected from the simulation environment, by placing the robot at a series of positions within the map, with different orientations under the view-field of a camera. An image is recorded once there is a change in the robot pose or camera pose. With this method, a dataset of 104,000 training images and 19,200 testing images was collected. To simulate the vibration and tilting of a flying drone, random noise is introduced into the camera pose, and this can also be regarded as a type of data augmentation that benefits CNN training. This data collection process is less-time consuming and cheap to validate the effectiveness of a proposed CNN on SCAMP-5d before performing testing in the real world.

\begin{figure}[t]
\centering
\includegraphics[width=2.8in]{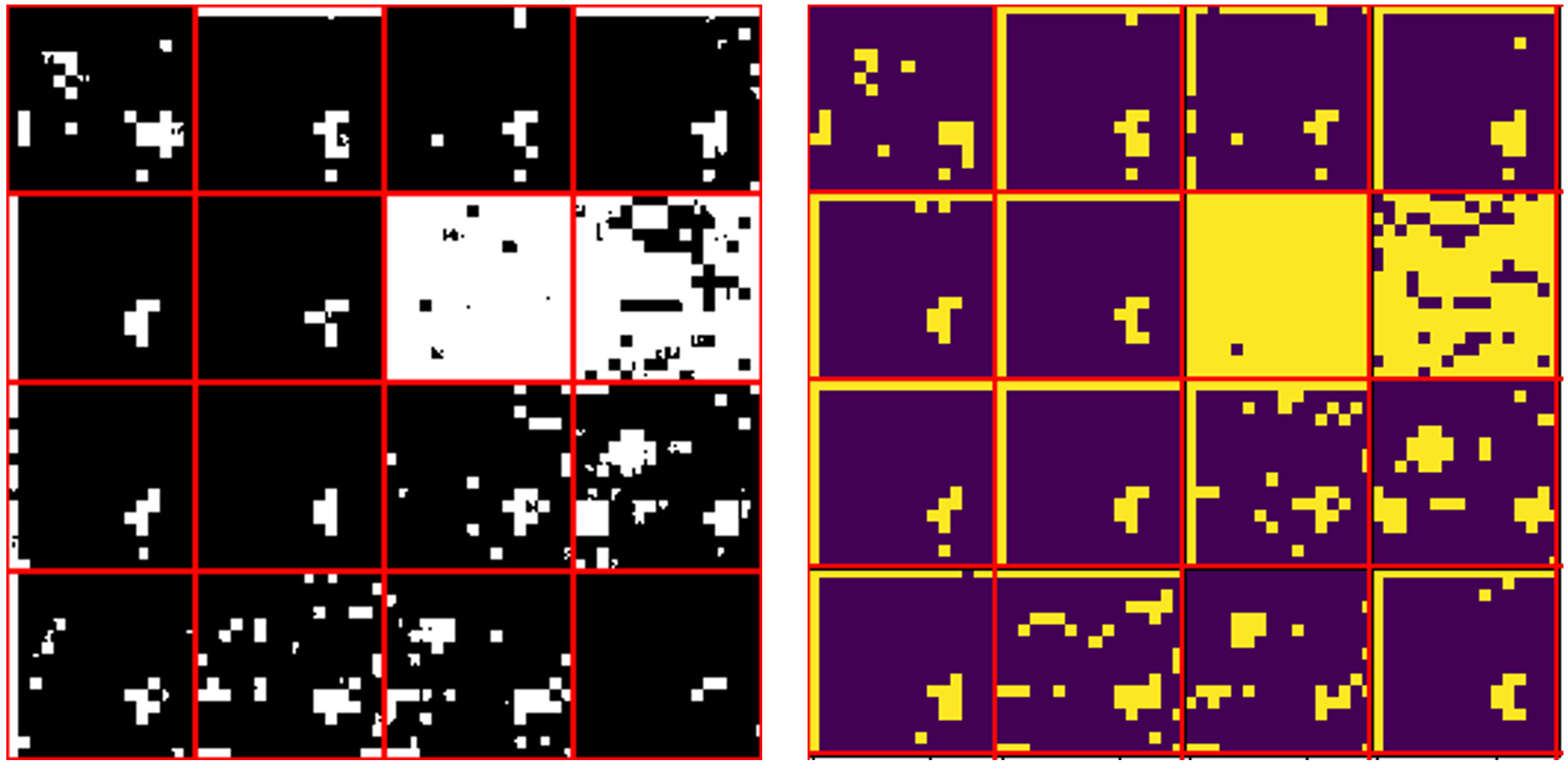}
\caption{Binary activations comparison after image convolution on SCAMP and PyTorch simulation. White and yellow dots represent '1's while the dark area is '-1's. This shows the similarity of binary activations after the first convolutional layer.}
\label{fig:cov_comparison}
\end{figure}

\begin{figure}[t]
\centering
\includegraphics[width=2.8in]{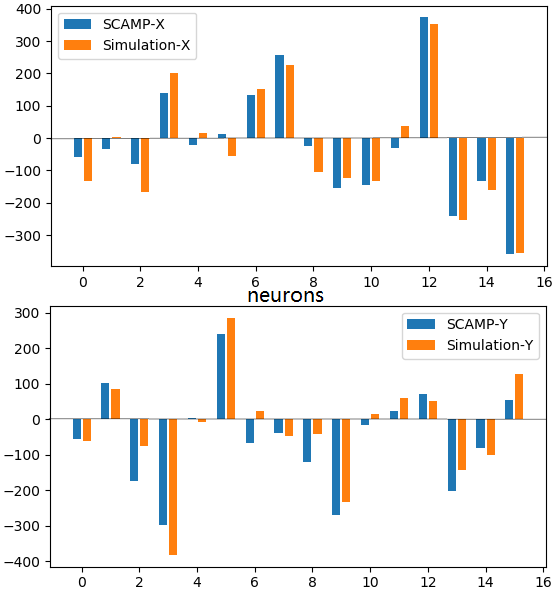}
\caption{Neuron activations after the first fully connected layer between SCAMP and PyTorch simulation. Due to binary activations and bit counting, we mitigate the error issue caused by using AREG for signal processing, hence getting a similar activation value from simulation to a real SCAMP.}
\label{fig:fc_comparison}
\end{figure}

\begin{figure}[t]
\centering
\includegraphics[width=3in]{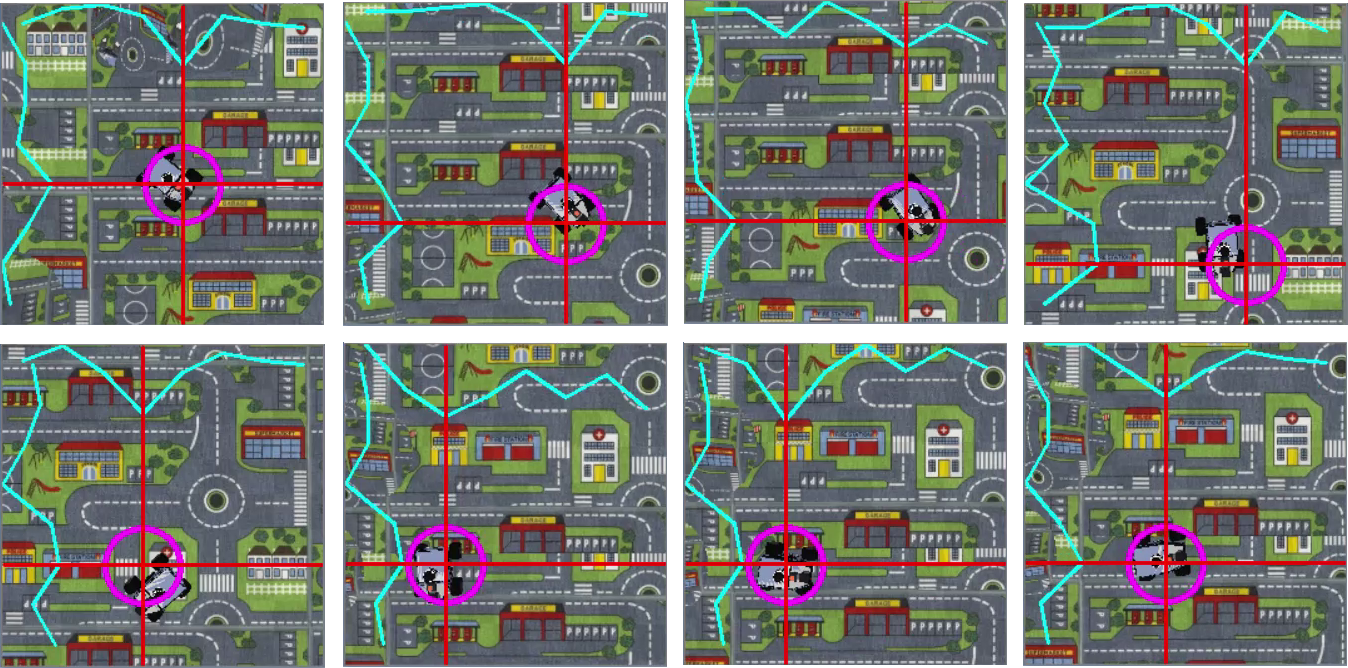}
\caption{Robot localisation result visualisation using SCAMP CNN inference results along with consecutive frames. The light blue prediction curves are plotted along \textit{x} and \textit{y} axis with possibilities for each label. With the largest possibilities for each axis (red straight line), the final localisation prediction (pink circle) is plotted. The full experimental video for robot localisation and tracking can be seen from \textit{\url{https://youtu.be/semthdfXH5A}}.}
\label{fig:robot_localisation} 
\end{figure}

\subsubsection{CNN inference for robot localisation}

\begin{figure}[t]
\centering
\includegraphics[width=3.0in]{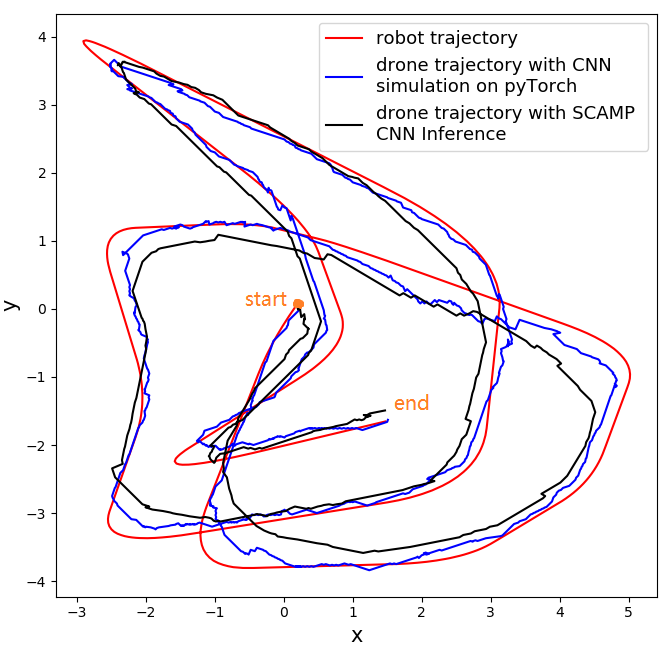}
\caption{Tracking trajectories with a drone. There are three paths: the pre-set robot trajectory as the groundtruth, drone tracking trajectory with CNN on PyTorch guidance, and drone tracking trajectory with guidance from SCAMP CNN inference.}
\label{fig:trackingtrajectory}
\end{figure}

This localisation task is regarded as a classification task by splitting the \textit{x} and the \textit{y} axis into eight labels, giving 64 possible positions to localise the robot. As for the CNN training, the training loss for backpropagation is the loss summation of \textit{x} and \textit{y} as shown in Fig. \ref{fig:cnn-localisation}. The final testing accuracy for localisation is around 93\%, which conservatively, only counts the correct predictions of the CNN. In a practical situation, a close prediction to the ground truth should still allow tracking to proceed without significant difficulty. Fig. \ref{fig:robot_localisation} visualises a sequence of 8 frames with CNN inference on SCAMP where the prediction possibility distribution can be seen from light blue curve along \textit{x} and \textit{y} axis, the final prediction is obtained with two highest possibilities from two curves along axes. The complete localisation and tracking are processed frame by frame, and the instructions to pilot the drone is generated using the PID control to minimise the distance between the robot ground truth position and the predicted robot position. 

To demonstrate the CNN inference on SCAMP in terms of accuracy, Fig. \ref{fig:cov_comparison} and Fig. \ref{fig:fc_comparison} compares the binary activation layer and first fully-connected layer between PyTorch and SCAMP, which shows a high similarity between the PyTorch simulation and SCAMP hardware implementation. The noise mainly results from the analogue signal information process, and it is inevitable due to the current method of manufacturing such hardware. However, this phenomenon could be effectively alleviated in the next generation of the SCAMP vision systems. To further validate the performance of CNN localisation on SCAMP, a chaotic trajectory is pre-set in the simulator for the robot to move along. The drone trajectory is plotted with guidance from SCAMP CNN inference. Fig. \ref{fig:trackingtrajectory} shows the comparison among the ground truth robot course, CNN on PyTorch guided drone course and CNN on SCAMP guided drone course. Finally, we make our SCAMP python host, CoppeliaSim model and its configuration available online: \textit{\url{https://github.com/yananliusdu/scamp5d_interface}}.

\section{Conclusion}

In this work, we proposed a semi-simulated platform where a real SCAMP interacts with the robot simulator via remote API for a rapid prototype validation. The SCAMP CNN inference results with the simulated sensor readings can instruct the motion of an agent in the proposed platform. More applications related to the SCAMP and robots integration can be explored based on the developed platform.

\bibliographystyle{IEEEtran}
\bibliography{ref}
\end{document}